\documentclass{article}
\usepackage{spconf,amsmath,graphicx,hyperref}
\usepackage{booktabs,multirow,hhline}
\usepackage{cleveref}
\usepackage{amssymb}
\usepackage{parskip}
\usepackage{tikz}
\usetikzlibrary{shapes.geometric, arrows, graphs, positioning, quotes, fit, decorations.pathreplacing}
\usepackage{rotating}

\title{RadarDINO: Distilling a Visual Foundation Model into mmWave Radar for Human Mesh Reconstruction}
\name{Furkan Mert Algan, Zhifan Ni, Eckehard Steinbach}
\address{Technical University of Munich\\
         School of Computation, Information and Technology\\
         Chair of Media Technology\\
         Munich Institute of Robotics and Machine Intelligence\\
         \{fmert.algan, zhifan.ni, eckehard.steinbach\}@tum.de}

\begin{document}
%
\maketitle
\begin{abstract}

Millimeter-wave (mmWave) radar offers a privacy-preserving and illumination-robust alternative for human mesh reconstruction (HMR). However, high-fidelity radar HMR remains challenging. Sparse point cloud-based methods suffer from information loss and missed detections, while raw radar tensor methods like RT-Mesh lack fine-grained semantics and explicit local body modeling. In this paper, we propose RadarDINO, a radar HMR framework that distills visual semantics from a DINOv3 teacher into 3D radar representations and introduces explicit geometric supervision. DINOv3 descriptors are transferred to radar voxels via calibrated RGB-radar geometry, providing voxel- and body-part-level supervision for part-aware radar representations. We further incorporate vertex- and joint-losses to directly constrain reconstructed geometry. On the M4Human dataset, our method outperforms the SOTA baseline across all three evaluation splits.

\end{abstract}
\section{Introduction}
\label{sec:intro}

3D human mesh reconstruction is crucial for human-centric applications. While recent vision-based methods~\cite{SMPLer-X,TokenHMR} achieve remarkable fidelity, optical cameras are sensitive to lighting variations and raise privacy concerns. In contrast, millimeter-wave (mmWave) radar provides a promising privacy-preserving, illumination-robust alternative for future Integrated Sensing and Communication (ISAC) systems.

Despite these advantages, high-fidelity radar HMR remains challenging. Early human pose estimation (HPE)~\cite{mRI,mm-Pose,MM-Fi,PoinTS} and human mesh reconstruction (HMR)~\cite{mmMesh,mmBody,M4esh} methods rely mainly on sparse radar point clouds (RPCs), where CFAR filtering discards informative radar responses and can introduce missed detections. Several works~\cite{HuPR,RT-Pose,RETR,RAPTR} instead exploit raw radar tensors (RTs) to preserve richer information, but remain focused on skeletal landmarks.

Recently, the M4Human~\cite{M4Human} benchmark introduced RT-Mesh, a raw-RT HMR baseline using 3D convolutions and self-attention. However, several limitations remain. First, raw RTs lack fine-grained semantics needed to distinguish body parts. Second, RT-Mesh does not explicitly model voxel- and part-level semantic correspondences, limiting localized body representation. Third, SMPL-X~\cite{SMPL-X} parametric supervision does not directly constrain reconstructed vertices and joints, which may lead to geometric errors under complex motions.

To address these challenges, we propose RadarDINO, a raw-radar HMR framework that combines cross-modal visual semantic distillation with explicit geometric supervision. Specifically, we transfer DINOv3~\cite{DINOv3} visual descriptors from synchronized RGB images to 3D radar voxels through calibrated radar-camera geometry, providing voxel- and body-part-level semantic supervision. The resulting part-aware radar representation is used for mesh reconstruction, while explicit vertex- and joint-level losses complement SMPL-X parameter supervision to directly constrain the reconstructed geometry. Extensive experiments on the M4Human dataset~\cite{M4Human} demonstrate that our model consistently improves radar-based HMR over existing methods, particularly under complex human motions.

\section{Methodology}
\begin{figure*}[ht]
\centering
\begin{tikzpicture}[font=\scriptsize, >={latex}]
    \node[inner sep=0] (root) {};
    \node[draw, cyan!50, rectangle, align=center, fill=cyan!10, thin, rounded corners=3pt, minimum height=195pt, minimum width=155pt, below right=18pt and -50pt of root] {};
    \node[draw, pink!100, rectangle, align=center, fill=pink!15, thin, rounded corners=3pt, minimum height=238pt, minimum width=158pt, below right=-25pt and 107pt of root] (frame_enc) {};
    \node[draw, green!40, rectangle, align=center, fill=green!10, thin, rounded corners=3pt, minimum height=238pt, minimum width=162pt, below right=-25pt and 273pt of root] {};

    \node[inner sep=0] (rgb_in) {\includegraphics[width=50px]{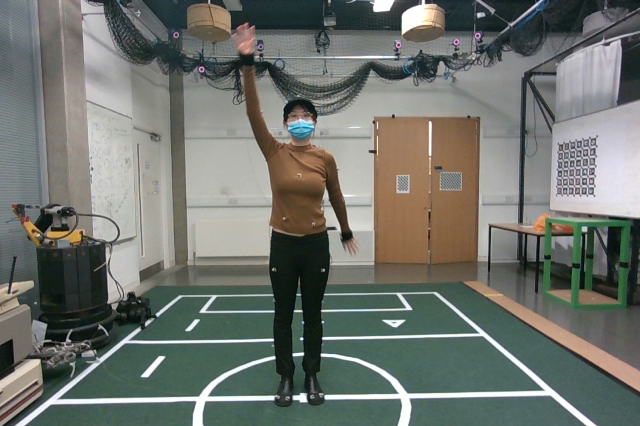}};
    \node[inner sep=0, right=25pt of rgb_in] (rt_in) {\includegraphics[width=50px]{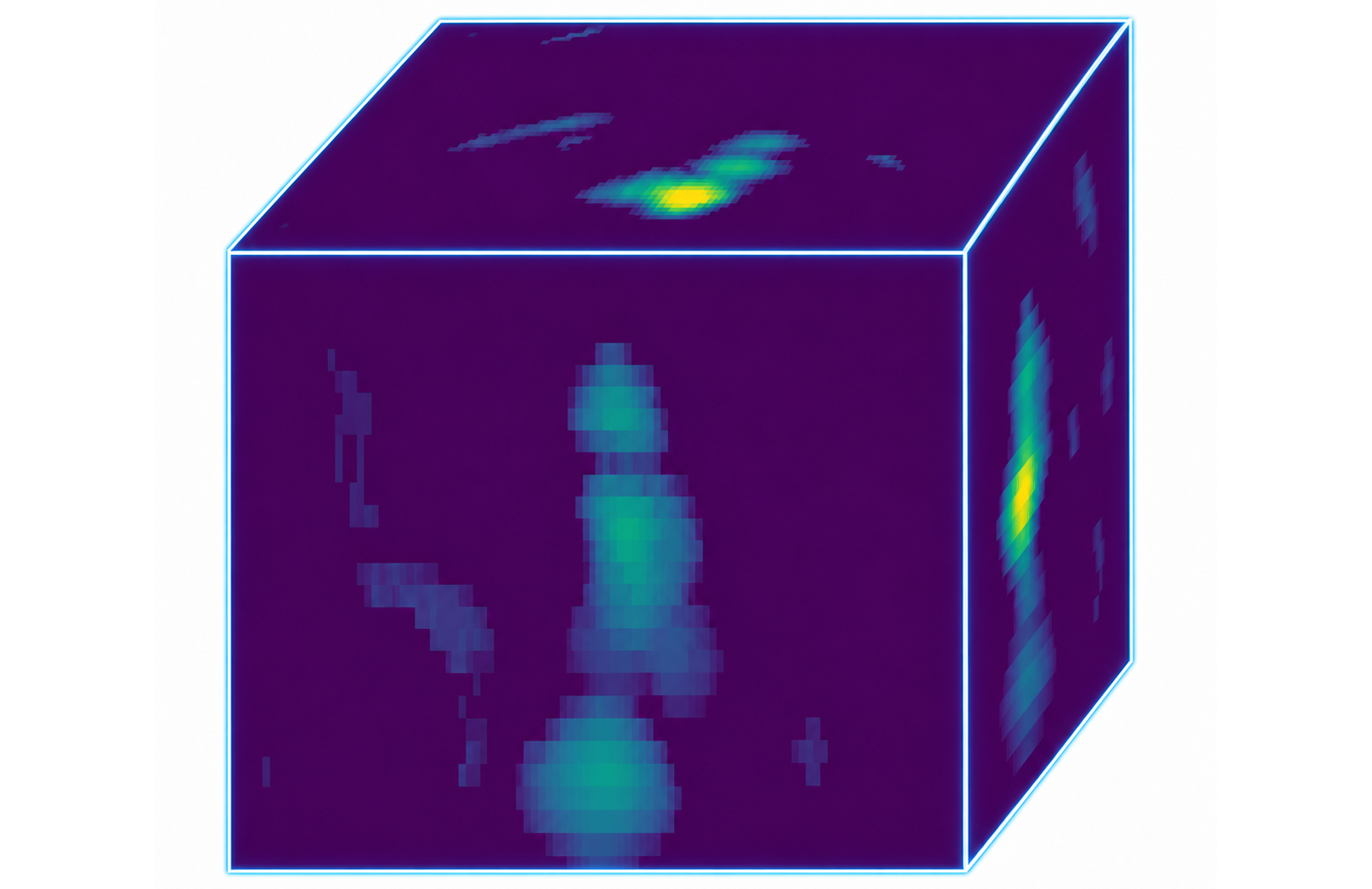}};
    \node[draw=none, rectangle, align=center, above=0pt of rgb_in] {RGB images $I_t$};
    \node[draw=none, rectangle, align=center, above=0pt of rt_in] {Radar tensors $X_t$};
    
    \node[draw, rectangle, align=center, fill=cyan!50, thin, rounded corners=3pt, minimum height=25pt, minimum width=30pt, below=7pt of rgb_in, xshift=-21pt] (dinov3) {\parbox[c]{30pt}{\centering DINOv3\\Encoder}};
    \node[draw, rectangle, align=center, fill=cyan!50, thin, rounded corners=3pt, minimum height=25pt, minimum width=30pt, below=7pt of rgb_in, xshift=21pt] (maskrcnn) {\parbox[c]{30pt}{\centering Mask\\R-CNN}};
    \node[draw, rectangle, align=center, fill=cyan!50, thin, rounded corners=3pt, minimum height=25pt, minimum width=40pt, below=7pt of rt_in] (proj) {\parbox[c]{40pt}{\centering Perspective\\Projection}};
    \node[draw=none, rectangle, align=center, below=2pt of dinov3] (Dt) {Feature map\\$D_t$};
    \node[draw=none, rectangle, align=center, below=2pt of maskrcnn] (Mt) {Person mask\\$M_t$};
    \node[draw, rectangle, align=center, fill=cyan!50, thin, rounded corners=3pt, minimum height=15pt, minimum width=80pt, below=5pt of Mt, xshift=-20pt] (foreground) {Foreground Selection};
    \node[draw=none, rectangle, align=center, below=5pt of foreground] (patch) {Human DINO patches};
    \node[inner sep=0, below=0pt of patch] (feat) {\includegraphics[width=50px]{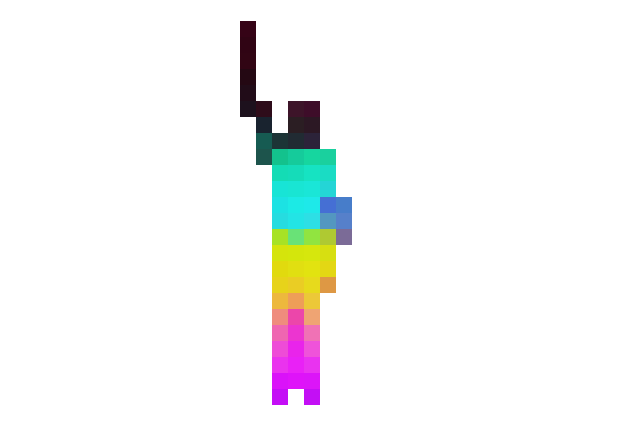}};
    \node[draw, rectangle, align=center, fill=cyan!50, thin, rounded corners=3pt, minimum height=25pt, minimum width=40pt, below=22pt of proj] (mapping) {\parbox[c]{40pt}{\centering Voxel-Patch\\Mapping}};
    \node[draw=none, rectangle, align=center, below=5pt of mapping] (pair) {Radar-visual\\correspondences\\$(\mathbf q_i,D_t[\mathbf p_i^D])$};
    \node[draw, rectangle, align=center, fill=cyan!50, thin, rounded corners=3pt, minimum height=15pt, minimum width=45pt, below right=20pt and -12pt of feat] (weighting) {\parbox[c]{45pt}{\centering Part-Specific\\Weighting}};
    \node[draw=none, rectangle, align=center, right=1pt of weighting, yshift=3pt] (part_teacher) {Body-part\\teacher\\$\mathbf d_{t,k}^{\mathrm{part}}$};
    \node[inner sep=0, left=7pt of weighting] (gt_weight) {\includegraphics[width=40px]{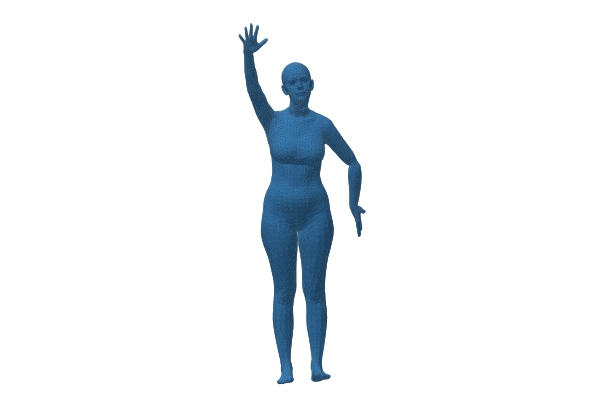}};
    \node[draw=none, rectangle, align=center, above=0pt of gt_weight] {Projected GT Mesh};
    \node[draw=none, rectangle, align=center, fill=cyan!30, rounded corners=3pt, minimum height=15pt, minimum width=130pt, below=8pt of weighting, xshift=-7pt] (banner_constr) {\textbf{Radar-Visual Teacher Construction}};
    \draw[->] ([xshift=-21pt]rgb_in.south) -> (dinov3.north);
    \draw[->] ([xshift=21pt]rgb_in.south) -> (maskrcnn.north);
    \draw[-] ([xshift=0pt]dinov3.south) -> ([yshift=-3pt]Dt.north);
    \draw[-] ([xshift=0pt]maskrcnn.south) -> ([yshift=-3pt]Mt.north);
    \draw[->] ([yshift=2pt]Dt.south) -> ([xshift=-22pt]foreground.north);
    \draw[->] ([yshift=2pt]Mt.south) -> ([xshift=20pt]foreground.north);
    \draw[->] (foreground.south) -> ([yshift=-3pt]patch.north);
    \draw[->] (rt_in.south) -> (proj.north);
    \draw[->] (proj.south) -> (mapping.north);
    \draw[->] (mapping.south) -> ([yshift=-3pt]pair.north);
    \draw[->] (feat.east) --++ (18pt, 0) |- ([yshift=-3pt]mapping.west);
    \draw[->] (feat.east) --++ (18pt, 0) -> ([xshift=4pt]weighting.north);
    \draw[->] (weighting.east) -> ([xshift=7pt, yshift=-3pt]part_teacher.west);
    \draw[->] (gt_weight.east) -> (weighting.west);

    \node[draw=none, rectangle, align=center, right=20pt of rt_in, yshift=-10pt] (fvoxel) {Voxel \\representation\\$\mathbf f_{t,i}$};
    \node[draw=none, rectangle, align=center, above left=7pt and -22pt of fvoxel] {$T$ frames};
    \node[draw, rectangle, align=center, fill=pink!100, thin, rounded corners=3pt, minimum height=25pt, minimum width=50pt, right=100pt of rt_in, yshift=-10pt] (cnn) {\parbox[c]{50pt}{\centering Multi-scale\\3D-CNN}};
    \node[draw, rectangle, align=center, fill=pink!100, thin, rounded corners=3pt, minimum height=25pt, minimum width=50pt, below=7pt of cnn] (stsa) {\parbox[c]{50pt}{\centering Spatiotemporal\\Self-Attention}};
    \node[draw=none, rectangle, align=center, below=4pt of stsa] (context) {Radar context\\$\mathbf F_t$};
    \node[draw, rectangle, align=center, fill=pink!100, thin, rounded corners=3pt, minimum height=30pt, minimum width=40pt, below=15pt of fvoxel] (ca) {\parbox[c]{40pt}{\centering Cross-\\Attention}};
    \node[draw=none, rectangle, align=center, above=0pt of ca, xshift=5pt] {Q};
    \node[draw=none, rectangle, align=center, right=0pt of ca, yshift=-5pt] {K,V};
    \node[draw, rectangle, align=center, fill=pink!100, thin, rounded corners=3pt, minimum height=25pt, minimum width=30pt, below=8pt of ca, xshift=-10pt] (desc_head) {\parbox[c]{30pt}{\centering Descriptor\\Head}};
    \node[draw, rectangle, align=center, fill=pink!100, thin, rounded corners=3pt, minimum height=25pt, minimum width=30pt, below=8pt of ca, xshift=40pt] (supp_head) {\parbox[c]{30pt}{\centering Support\\Head}};
    \node[draw=none, rectangle, align=center, below=4pt of desc_head] (dvoxel) {Voxel descriptor\\$\hat{\mathbf d}_{t,i}$};
    \node[draw=none, rectangle, align=center, below=4pt of supp_head] (supp) {Support logit\\$s_{t,i}$};
    \node[draw, rectangle, align=center, fill=pink!100, thin, rounded corners=3pt, minimum height=30pt, minimum width=30pt, below=55pt of context] (ca_mlp) {\parbox[c]{50pt}{\centering Cross-Attention\\+ MLP Head}};
    \node[draw, rectangle, align=center, fill=pink!100, thin, rounded corners=17pt, minimum height=34pt, minimum width=40pt, left=15pt of ca_mlp] (bodyQ) {\parbox[c]{40pt}{\centering Learned\\Body-Part\\Queries}};
    \node[draw=none, rectangle, align=center, left=1pt of ca_mlp, yshift=5pt] {Q};
    \node[draw=none, rectangle, align=center, above=0pt of ca_mlp, xshift=9pt] {K,V};
    \node[draw=none, rectangle, align=center, below=5pt of ca_mlp] (dpart) {Part descriptor $\hat{\mathbf d}_{t,k}^{\mathrm{part}}$};
    \node[draw=none, rectangle, align=center, fill=pink!50, rounded corners=3pt, minimum height=15pt, minimum width=130pt, right=28pt of banner_constr] (banner_enc) {\textbf{Distilled Radar Encoder}};
    \draw[->] ([yshift=13pt]rt_in.east) -| ([yshift=-3pt]fvoxel.north);
    \draw[->] ([yshift=13pt]rt_in.east) -| (cnn.north);
    \draw[->] (fvoxel.south) -> (ca.north);
    \draw[->] (cnn.south) -> (stsa.north);
    \draw[-] (stsa.south) -> ([yshift=-3pt]context.north);
    \draw[->] ([yshift=4.5pt]context.west) -> ([yshift=-10pt]ca.east);
    \draw[->] (bodyQ.east) -> (ca_mlp.west);
    \draw[->] (context.south) -> (ca_mlp.north);
    \draw[->] ([xshift=-10pt]ca.south) -> (desc_head.north);
    \draw[->] ([xshift=10pt]ca.south) --++ (0, -2.5pt) -| (supp_head.north);
    \draw[->] (desc_head.south) -> ([yshift=-4pt]dvoxel.north);
    \draw[->] (supp_head.south) -> ([yshift=-4pt]supp.north);
    \draw[->] (ca_mlp.south) -> ([yshift=-4pt]dpart.north);

    \node[draw, trapezium, align=center, fill=pink!15, thin, anchor=north, rotate=180, minimum height=30pt, trapezium stretches body, right=250pt of rt_in, yshift=12pt] (enc) {\rotatebox{180}{\parbox[c]{30pt}{\centering Distilled\\Radar\\Encoder}}};
    \node[draw=none, rectangle, align=center, below=33pt of enc] (enc_out) {$\hat{\mathbf d}_{t,i}, \hat{\mathbf d}_{t,k}^{\mathrm{part}}, s_{t,i}$};
    \node[draw, trapezium, align=center, fill=green!50, thin, anchor=north, rotate=0, minimum height=25pt, trapezium stretches body, below=15pt of enc_out] (dec) {\rotatebox{0}{\parbox[c]{30pt}{\centering Mesh\\Decoder}}};
    \node[draw=none, rectangle, align=center, above=0pt of dec, xshift=30pt] {+ voxel coordinates \\ and intensity};
    \node[draw=none, rectangle, align=center, below=7pt of dec] (smplx) {SMPL-X params};
    \node[draw, rectangle, align=center, fill=green!50, thin, rounded corners=3pt, minimum height=25pt, minimum width=50pt, below=7pt of smplx] (smpl_layer) {\parbox[c]{50pt}{\centering Frozen\\SMPL-X Layer}};
    \node[draw=none, rectangle, align=center, below=6pt of smpl_layer] (pred_text) {Mesh prediction};
    \node[draw=none, rectangle, align=center, below=-4pt of pred_text] (pred) {\includegraphics[width=45px]{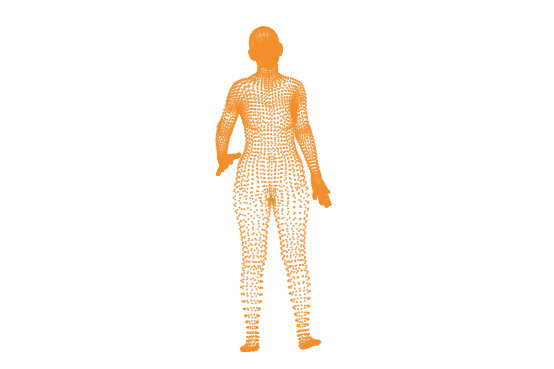}};
    \node[draw=none, rectangle, align=center, right=35pt of pred_text] (gt_text) {GT Mesh};
    \node[draw=none, rectangle, align=center, below=-3.5pt of gt_text] (gt_mesh) {\includegraphics[width=45px]{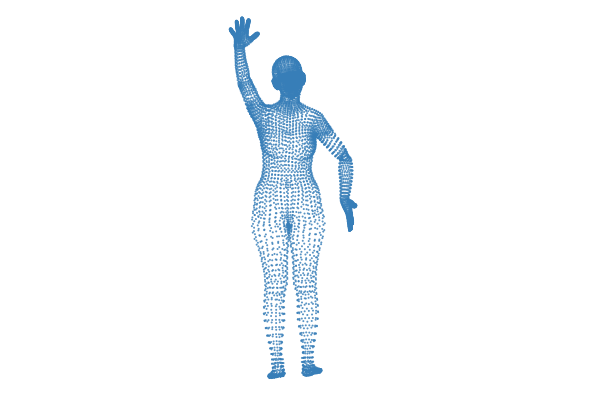}};
    \node[draw=none, rectangle, align=center, above=40pt of gt_text] (gt_params) {GT Params};
    \node[draw=none, rectangle, align=center, fill=green!25, rounded corners=3pt, minimum height=15pt, minimum width=130pt, right=32pt of banner_enc] {\textbf{Human Mesh Reconstruction}};
    
    \draw[->] ([yshift=13pt]rt_in.east) -| (enc.south);
    \draw[-] (enc.north) -> ([yshift=-2pt]enc_out.north);
    \draw[->] ([yshift=3pt]enc_out.south) -> (dec.north);
    \draw[->] (dec.south) -> ([yshift=-2pt]smplx.north);
    \draw[->] ([yshift=3pt]smplx.south) -> (smpl_layer.north);
    \draw[->] (smpl_layer.south) -> ([yshift=-2pt]pred_text.north);
    \draw[->] (gt_text.north) -> (gt_params.south);

    \node[draw=none, rectangle, align=center, below right=-1pt and 5pt of pair] {\textcolor{red}{$\mathcal L_{\mathrm{voxel}}$}};
    \draw[->, dashed, red] (pair.south) --++ (0, -10pt) -| (dvoxel.south);
    \node[draw=none, rectangle, align=center, above right=-25pt and 11pt of part_teacher] {\textcolor{red}{$\mathcal L_{\mathrm{part}}$}};
    \draw[->, dashed, red] ([yshift=-7pt]part_teacher.east) -> (dpart.west);
    
    \node[draw=none, rectangle, align=center, above left=-21pt and 1pt of gt_mesh] {\textcolor{red}{$\mathcal L_{\mathrm{geo}}$}};
    \draw[->, dashed, red] (gt_mesh.west) -> (pred.east);
    \node[draw=none, rectangle, align=center, above left=-8pt and 0pt of gt_params] {\textcolor{red}{$\mathcal L_{\mathrm{param}}$}};
    \draw[->, dashed, red] (gt_params.west) -> (smplx.east);

    \draw[->, pink!100, very thick] (frame_enc.east) --++ (15pt, 0) |- (enc.east);
\end{tikzpicture}
\caption{The radar-visual distillation pipeline and the overall HMR pipeline of our proposed RadarDINO.}
\label{fig:pipeline}
\end{figure*}

As illustrated in Fig.~\ref{fig:pipeline}, our framework follows a two-stage pipeline. We first distill DINOv3~\cite{DINOv3} semantics into a spatio-temporal radar encoder using sparse voxel correspondences and body-part-level supervision. A mesh decoder then uses distilled representation to predict SMPL-X parameters and reconstruct the human geometry from radar input only.

\subsection{Radar-Visual Teacher Construction}
\label{sec:dino_preprocessing}

Following prior DINO distillation methods~\cite{DINOintheRoom,DinoComplete}, we construct sparse visual teacher features at 3D radar locations using synchronized RGB images and calibrated radar-camera geometry. A frozen DINOv3 ViT-S/16 encoder~\cite{DINOv3} extracts a feature map $D_t\in\mathbb R^{H_p\times W_p\times C}$ from each $H\times W$ RGB image. A pretrained Mask R-CNN~\cite{maskrcnn} detects the person, and the largest valid instance is retained. A DINO patch is marked as foreground if at least $40\%$ of its corresponding pixels belong to the person mask, yielding $M_t\in\{0,1\}^{H_p\times W_p}$.

We associate DINO descriptors with radar voxels via geometric projection. Following M4Human~\cite{M4Human}, a voxel with grid coordinate $\mathbf q_i$ in an RT is mapped to its metric position by
\begin{equation}
    \mathbf p_i^r =
    \mathbf b_{\min}
    + \frac{\mathbf q_i}{\mathbf N}\odot
    (\mathbf b_{\max}-\mathbf b_{\min}),
\end{equation}
where $\mathbf N=(121,111,31)^{\mathsf T}$ denotes the RT-grid size in the M4Human dataset~\cite{M4Human}, $\mathbf b_{\min}$ and $\mathbf b_{\max}$ represent the metric spatial bounds. The voxel is transformed into the camera frame and projected onto the image plane using the extrinsic transformation $(R_{c\leftarrow r}, \mathbf t_{c\leftarrow r})$ and the camera intrinsics $K$:
\begin{equation}
    \mathbf p_i^c=R_{c\leftarrow r}\mathbf p_i^r+\mathbf t_{c\leftarrow r},
    \quad
    (u_i,v_i)=\pi(K\mathbf p_i^c),
\end{equation}
where $\pi(\cdot)$ is the perspective projection. The position of the corresponding DINO patch is
\begin{equation}
    \mathbf p_i^D=
    \left(
    \left\lfloor\frac{v_iH_p}{H}\right\rfloor,
    \left\lfloor\frac{u_iW_p}{W}\right\rfloor
    \right).
\end{equation}
Voxels that lie in front of the camera, project inside the image, and fall on foreground patches are paired with the corresponding descriptor $D_t[\mathbf p_i^D]$.


For offline teacher construction, we center the radar crop using M4Human's ground-truth horizontal root position transformed into the radar frame. Within the crop, we retain up to three strongest positive-valued voxels per foreground patch and at most $3000$ foreground voxels per frame. The resulting pairs $(\mathbf q_i,D_t[\mathbf p_i^D])$ form the sparse radar-visual teacher correspondences used for DINO distillation.

\subsection{DINO Distillation into Radar Features}
\label{sec:dino_distillation}

We train the radar backbone to reproduce the visual semantics provided by the constructed sparse teacher correspondences. Following M4Human~\cite{M4Human}, the input is a sequence of four radar tensors, $\mathcal X_t=[X_{t-3},\ldots,X_t]$, where $X_t\in\mathbb R^{121\times 111\times 31}$.

A localization branch first predicts the horizontal root position
$\hat{\mathbf p}_t^{\mathrm{root}}\in\mathbb R^2$ from the full radar sequence. Following the M4Human cropping convention, this prediction defines a spatial crop shared across all four frames. A shared 3D CNN extracts multi-scale features from each cropped frame. Then, spatial and temporal self-attentions aggregate them into a shared radar context $\mathbf F_t$.

For each selected voxel $\mathbf q_i$ within the crop, we construct its representation $\mathbf f_{t,i} = [\mathbf c_{t,i}; \mathbf x_{t,i}; \tilde{\mathbf q}_{t,i}^{\mathrm{local}}; \tilde{\mathbf q}_{t,i}^{\mathrm{global}}]$, where $\mathbf c_{t,i}$ contains multi-scale CNN features sampled at the voxel location, $\mathbf x_{t,i}$ contains its radar values across the input frames, and the last two terms encode its normalized local and global coordinates. The voxel representation then queries $\mathbf F_t$ via cross-attention, followed by MLP prediction heads that output an $\ell_2$-normalized DINO descriptor $\hat{\mathbf d}_{t,i}$ and a person-support logit $s_{t,i}$, which estimates whether the voxel corresponds to a human body and is later used in the mesh decoder.

In parallel, learned body queries attend to the same radar context and produce semantic descriptors $\hat{\mathbf d}_{t,k}^{\mathrm{part}}$ for the whole body ($k=0$) and seven body parts ($k\in\{1,\dots,7\}$).(instead of 22 joints to reduce the number of descriptors) These part-aware queries encourage the radar representations to preserve local body semantics and dependencies.

The voxel descriptors $\hat{\mathbf d}_{t,i}$ are supervised by the precomputed DINO teacher features $D_t[\mathbf p_i^D]$. Let $\mathcal V_t$ denote the valid foreground voxels within the predicted crop. We minimize the cosine distance
\begin{equation}
    \mathcal L_{\mathrm{voxel}} = \frac{\sum_t\sum_{i\in\mathcal V_t}w_{t,i}\left[1-\cos\left(\hat{\mathbf d}_{t,i},D_t[\mathbf p_i^D]\right)\right]}{\sum_t\sum_{i\in\mathcal V_t} w_{t,i}},
\end{equation}
where $w_{t,i}=1/n_{t,p_i}$ and $n_{t,p}$ is the number of selected radar voxels associated with patch $p$. This weighting prevents image patches receiving multiple radar correspondences from dominating the distillation objective.

The support head is trained with class-balanced binary cross-entropy $\mathcal L_{\mathrm{support}}$ using the projected person-masks $M_t[\mathbf p_i^D]$. The localization branch is supervised with a Smooth-$L_1$ loss $\mathcal L_{\mathrm{center}}$ on the ground-truth horizontal root position.

To provide structured semantic supervision beyond individual voxels, we construct DINO teachers for seven body parts. For each part $k$, we compute a weighted average of foreground patch descriptors using weights derived from the projected ground-truth SMPL-X mesh, and normalize the result to obtain $\mathbf d_{t,k}^{\mathrm{part}}$. Valid body-part descriptors are supervised against these teachers using a cosine distance $\mathcal L_{\mathrm{part}}$. The complete distillation objective is a weighted combination of the voxel-level distillation, support, body-part distillation, and localization losses.

\subsection{Mesh Prediction}
\label{sec:mesh_prediction}

We initialize the radar backbone with the distilled model and fine-tune it jointly with a mesh decoder. Let $\mathcal S_t$ denote the selected positive-valued voxels within the predicted crop. For each voxel $i\in\mathcal S_t$, we construct a mesh feature by combining its distilled semantic descriptor, spatial information, radar intensity, and person-support score:
\begin{equation}
    \mathbf m_{t,i} = \sigma(s_{t,i})\psi\left([\hat{\mathbf d}_{t,i};\tilde{\mathbf q}_{t,i}^{\mathrm{local}};\mathbf p_i^r-\mathbf o_t;\rho_{t,i}]\right),
\end{equation}
where $\psi(\cdot)$ is a learned projection, $\rho_{t,i}$ denotes the normalized log radar intensity, and
$\mathbf o_t=[\hat{\mathbf p}_t^{\mathrm{root}};z_0]$
defines the 3D crop reference by extending the predicted horizontal root position with the fixed vertical crop center $z_0$. The person-support probability $\sigma(s_{t,i})$ suppresses voxels unlikely to belong to a human body.

The mesh decoder receives learned joint queries conditioned on the corresponding distilled body-part representations and a global query summarizing whole-body information. All queries also encode the predicted root and crop position. Collecting them as $\mathbf Q_t$, the decoder computes
\begin{equation}
    \mathbf Z_t = \operatorname{Decoder}\left(\mathbf Q_t,[\operatorname{Proj}(\mathbf F_t);\{\mathbf m_{t,i}\}_{i\in\mathcal S_t}]\right),
\end{equation}
where $\operatorname{Proj}$ maps $\mathbf F_t$ to the decoder feature dimension.

The joint outputs in $\mathbf Z_t$ predict the root and body rotations $\hat{\boldsymbol\Theta}_t$, while the global output predicts the body shape $\hat{\boldsymbol\beta}_t$, a translation residual $\Delta\hat{\mathbf t}_t$, and a gender logit $\hat g_t$. The final translation is $\hat{\mathbf t}_t = \mathbf o_t+\Delta\hat{\mathbf t}_t$.

A fixed SMPL-X layer then converts
$(\hat{\boldsymbol\Theta}_t,\hat{\boldsymbol\beta}_t,\hat{\mathbf t}_t)$
into mesh vertices $\hat{\mathbf V}_t$ and articulated joints
$\hat{\mathbf J}_t$. Ground-truth gender is used during training, while inference selects the body model using the predicted gender and requires only radar input.

Following M4Human~\cite{M4Human}, we supervise the predicted SMPL-X parameters using shape, pose, root-orientation, translation, and gender losses. We further introduce a geometry loss $\mathcal L_{\mathrm{geo}}$, which combines MAE supervision on the reconstructed vertices and joints. The final mesh reconstruction objective is a weighted combination of the parameter, geometry, and localization losses.

\begin{table*}[ht]
\centering

\renewcommand{\arraystretch}{1.1}
\resizebox{\textwidth}{!}{
\begin{tabular}{c|c|cc|cccc|cccc|cccc}
\hline
\multirow{2}{*}{Modality} &
\multirow{2}{*}{Method} &
\multicolumn{2}{c|}{Efficiency} &
\multicolumn{4}{c|}{S1 (Random)} &
\multicolumn{4}{c|}{S2 (Cross-Subject)} &
\multicolumn{4}{c}{S3 (Cross-Action)} \\

\cline{3-16}

&
&
Param. (M) & GFLOPs &
MVE & MJE & MRE & TE &
MVE & MJE & MRE & TE &
MVE & MJE & MRE & TE \\
\hline

\multirow{2}{*}{RPC}
& mmMesh~\cite{mmMesh}
& 41.45 & 2.87
& 132.7 & 112.1 & 11.8 & 70.4
& 170.1 & 147.9 & 16.0 & 100.0
& 173.8 & 146.9 & 16.4 & 91.8 \\

& P4Trans.~\cite{Point4DTransformer}
& 129.01 & 11.76
& 89.5 & 76.6 & 9.2 & 48.6
& 140.8 & 125.2 & 15.0 & 86.4
& 147.8 & 126.4 & 16.1 & 78.4 \\
\hline

\multirow{4}{*}{RT}
& RT-Pose~\cite{RT-Pose}
& 6.05 & 50.67
& 100.7 & 87.0 & 9.9 & 56.7
& 148.1 & 131.2 & 15.3 & 93.9
& 152.8 & 131.0 & 16.0 & 84.6 \\

& RETR~\cite{RETR}
& 52.42 & 3.01
& 97.1 & 81.8 & 9.1 & 50.4
& 169.7 & 148.6 & 16.3 & 100.1
& 163.1 & 138.3 & 16.6 & 85.4 \\

& RT-Mesh~\cite{M4Human}
& 63.25 & 2.60
& 90.9 & 77.2 & 9.6 & 47.6
& 135.1 & 120.2 & 14.9 & 86.1
& 143.1 & 122.0 & 15.6 & 76.0 \\
\cline{2-16}

& \textbf{Ours}
& \textbf{2.70} & \textbf{1.98}
& \textbf{62.5} & \textbf{52.6} & \textbf{9.0} & \textbf{33.3}
& \textbf{120.0} & \textbf{108.2} & \textbf{14.2} & \textbf{82.2}
& \textbf{111.6} & \textbf{94.6} & \textbf{14.9} & \textbf{61.9} \\
\hhline{================}

RGB + RT & RT-Mesh~\cite{M4Human}
& -- & --
& 80.1 & 71.4 & 8.9 & 51.6
& 112.5 & 103.2 & 11.6 & 80.5
& 108.7 & 98.4 & 11.5 & 75.7 \\

\hline
\end{tabular}
}

\caption{Comparison of radar-based HMR methods on M4Human~\cite{M4Human} under the ALL protocol.
We report MVE, MJE, MRE, and TE under three data splits, together with model parameters and computational cost. Lower is better for all metrics.}
\label{tab:m4human_comparison}
\end{table*}

\section{Experiment}
\subsection{Experimental Setup}
We conduct experiments on the M4Human dataset~\cite{M4Human}, the current largest multimodal benchmark for mmWave radar-based HMR. We adhere to the official experimental protocols and dataset splits. We train the radar student for $25$ epochs and jointly fine-tune it with the mesh decoder for $20$ epochs, using AdamW with learning rates of $4\times10^{-4}$ and $2\times10^{-4}$, respectively, and cosine decay. Model performance is evaluated using standard metrics: Mean Vertex Error (MVE, mm) across all SMPL-X vertices, Mean Joint Error (MJE, mm) on $22$ body joints, Mean Rotation Error (MRE, $^\circ$) evaluating joint rotations on $SO(3)$, and Translation Error (TE, mm) measuring global root localization. The baseline results are taken from~\cite{M4Human}.

\subsection{Experimental Results}

\begin{figure}[ht]
    \centering
    \includegraphics[width=0.99\linewidth]{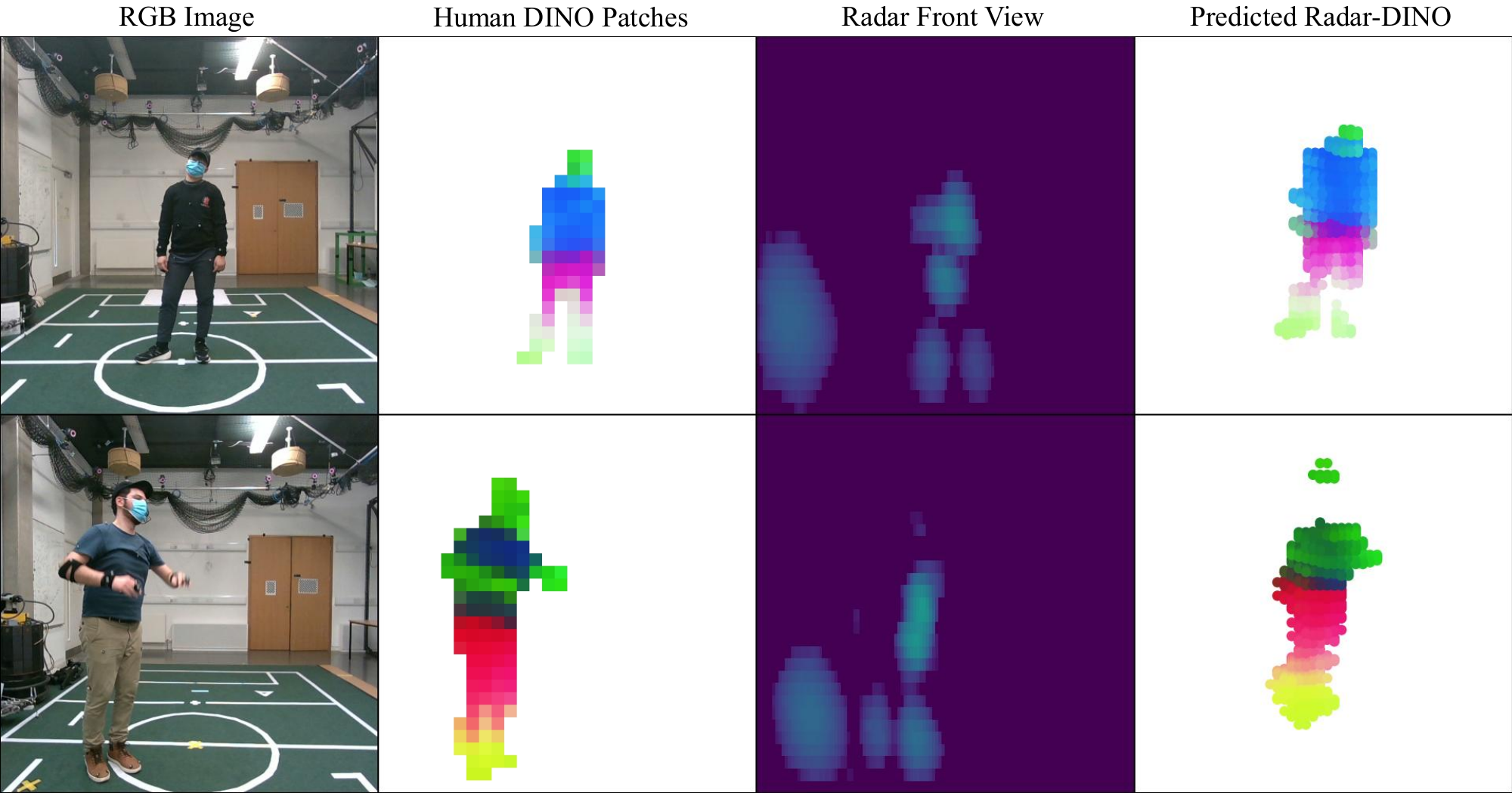}
    \caption{Visualization of distilled radar DINO features.}
    \label{fig:feature}
\end{figure}

\begin{figure}[ht]
    \centering
    \includegraphics[width=0.99\linewidth]{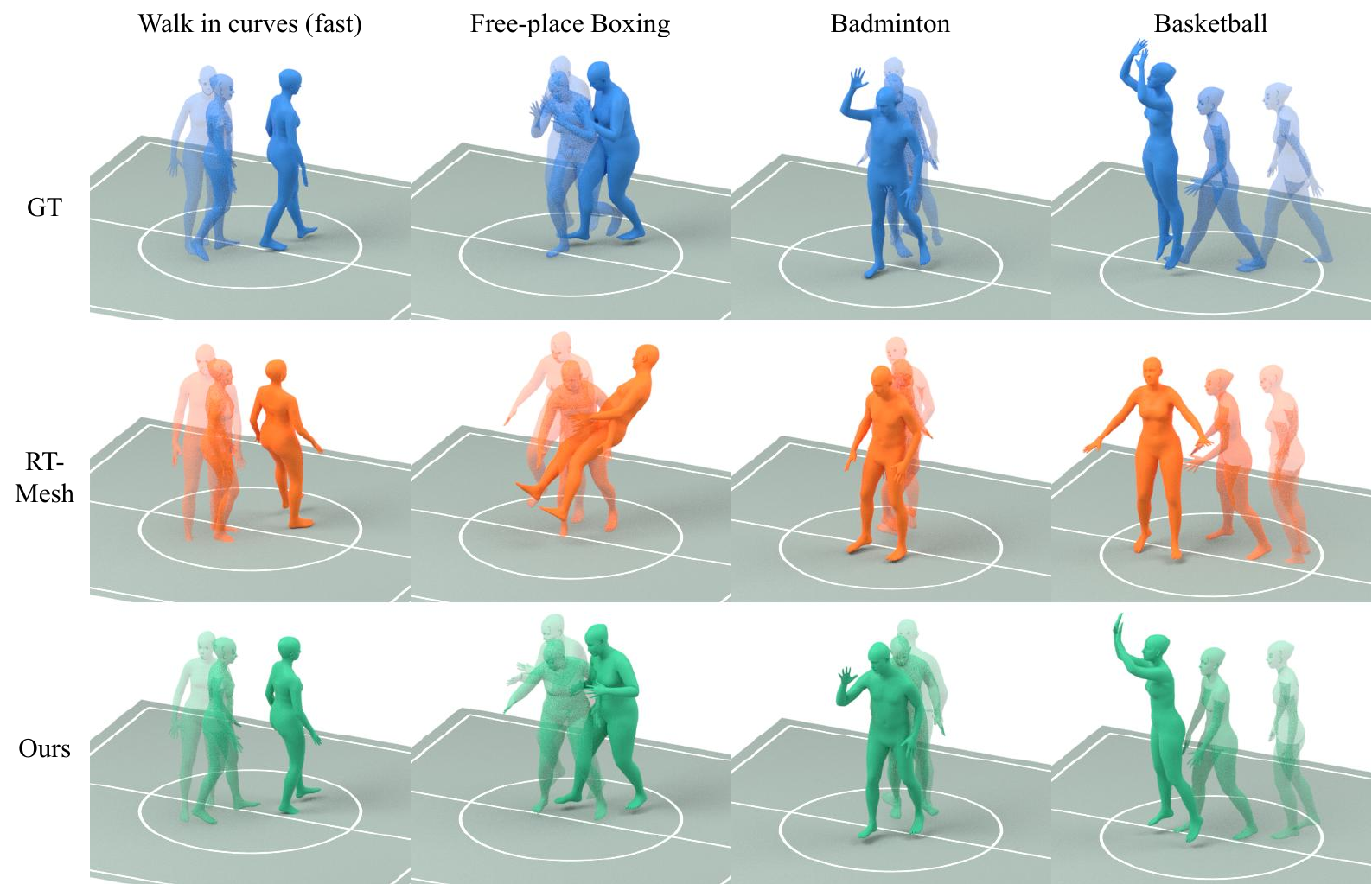}
    \caption{Visualization of RT HMR under non-in-place actions.}
    \label{fig:qualitative}
\end{figure}

\textbf{Quantitative Evaluation.}
\Cref{tab:m4human_comparison} compares our method with reported radar-based baselines on the three M4Human splits. Our method achieves the lowest MVE across all splits, reducing RT-Mesh's~\cite{M4Human} MVE by 31.3\%, 11.2\%, and 22.0\% on S1--S3, respectively, while also improving joint, rotation, and translation errors. Compared with RGB+RT fusion, our radar-only model remains competitive while requiring no RGB input at inference. In terms of efficiency, our model uses only 2.70M parameters while requiring 1.98 GFLOPs, achieving both the smallest model size and the lowest computational cost among RT only methods.



\textbf{Qualitative Evaluation. }
\Cref{fig:feature} compares the human DINO patch descriptors with their corresponding predicted radar representations. The radar-predicted descriptors largely preserve the body-part structure of the DINO teacher. Some local details are lost, likely due to the limited spatial resolution of the radar input, such as around the neck in the second example. Overall, the predicted descriptors remain visually similar to the teacher features across major body regions.


\Cref{fig:qualitative} compares reconstructed meshes in challenging non-in-place sequences, with earlier poses shown at reduced opacity for temporal context. Compared with RT-Mesh~\cite{M4Human}, our method produces more consistent limb configurations during fast curved walking and boxing, better captures raised-arm poses in sports sequences, and more closely matches the ground-truth body orientation.



\begin{table}[ht]
\centering
\resizebox{0.48\textwidth}{!}{
    \begin{tabular}{l|cccc}
    \hline
        Radar Encoder Weights & MVE & MJE & MRE & TE\\
    \hline
        Random initialization & 120.65 & 102.27 & 15.38 & 63.31\\
        DINO-distilled, frozen & 114.95 & 98.30 & 15.04 & 62.75\\
        DINO-distilled, fine-tuned & 111.58 & 94.62 & 14.90 & 61.87\\
    \hline
        RGB DINO teacher & 87.03 & 71.71 & 13.16 & 36.23\\
    \hline
    \end{tabular}
}
\caption{Ablation study on radar encoder initialization and fine-tuning strategies under the most challenging S3 split.}
\label{tab:ablation}
\end{table}
\subsection{Ablation Study}
\Cref{tab:ablation} evaluates DINO-guided radar pretraining on the most challenging S3 split. Compared with random initialization, DINO-guided initialization consistently improves mesh reconstruction. The gain is largely retained with a frozen encoder and further improves with joint fine-tuning, indicating that the distilled representation itself provides useful information for downstream mesh prediction.  However, directly using the RGB DINO teacher features achieves a much lower MVE than using the distilled radar representation on S3, revealing a remaining teacher-student gap.


\section{Conclusion}
We present RadarDINO, a two-stage framework for radar-based human mesh reconstruction via visual feature distillation. A fine-tuned visual foundation model supervises the radar encoder at voxel and body-part levels, and the distilled representation is used by a mesh decoder to predict SMPL-X geometry from radar alone. Experiments on M4Human show consistent improvements over existing radar-based baselines. Future work will explore more effective cross-modal distillation and stronger generalization across environments and radar devices.


\vfill\pagebreak
\section{Acknowledgements}
This work has been supported by the German Federal Ministry of Research, Technology and Space (BMFTR) under the Robotics Institute Germany (RIG) and the Bavarian Ministry of Economic Affairs, Regional Development and Energy as part of the project ``6G Future Lab Bavaria''.

OpenAI ChatGPT and Codex were used for language editing and to assist with code development and visualization.

\bibliographystyle{IEEEbib}
\bibliography{refs}

\end{document}